\documentclass[runningheads]{llncs}

\usepackage{eccv}

\usepackage{eccvabbrv}
\usepackage{colortbl}
\usepackage{array}
\usepackage{amsmath}
\usepackage{multirow}
\usepackage{makecell}
\usepackage{graphicx}
\usepackage{booktabs}
\usepackage{algorithmicx,algorithm}
\usepackage[noend]{algpseudocode}
\usepackage{color}
\usepackage{xcolor}
\usepackage{svg}
\usepackage{bbding}
\usepackage{caption}
\usepackage{wrapfig}
\usepackage{float}
\usepackage{graphicx}
\usepackage{svg} 
\usepackage{footnote}
\usepackage{footmisc}

\usepackage[accsupp]{axessibility}  

\usepackage[pagebackref,breaklinks,colorlinks,citecolor=eccvblue]{hyperref}
\usepackage{orcidlink}
\usepackage{enumitem}
\definecolor{cvprblue}{rgb}{0.21,0.49,0.74}
\definecolor{boost}{RGB}{243,220,205}
\definecolor{adapt}{RGB}{232,243,225}
\definecolor{boost_line}{RGB}{197,90,17}
\definecolor{adapt_line}{RGB}{84,130,53} 

\definecolor{tablecolor}{gray}{.93}
\definecolor{decline_color}{RGB}{75,140,174}

\newcommand{\implus}[1]{\textcolor{black}{#1}}
\newcommand{\rewrite}[1]{\textcolor{black}{#1}}
\newcommand{\important}[1]{\textcolor{black}{#1}}
\newcommand{\zhangxin}[1]{\textcolor{black}{#1}}
\newcommand{\hlboost}[1]{\colorbox{boost}{\raisebox{0pt}[5pt][0pt]{#1}}}
\newcommand{\hladapt}[1]{\colorbox{adapt}{\raisebox{0pt}[5pt][0pt]{#1}}}
\newcommand{\boostbox}[0]{\raisebox{0.25em}{\fcolorbox{boost_line}{boost}{}}}
\newcommand{\adaptbox}[0]{\raisebox{0.25em}{\fcolorbox{adapt_line}{adapt}{}}}
\newcommand{\indexbf}[1]{\textcolor{orange}{\textbf{#1}}}
\newcommand{\improve}[1]{\textcolor{boost_line}{\textbf{#1}}}
\newcommand{\decline}[1]{\textcolor{decline_color}{\textbf{#1}}}
\newcommand{\lz}[1]{\textcolor{black}{#1}}

\newcommand{\myfnsymbol}[1]{%
  \expandafter\myfnsymbola\csname c@#1\endcsname
}
\newcommand{\myfnsymbola}[1]{%
  \ifcase #1
  \or \TextOrMath{\textdagger}{\dagger}
  \or \TextOrMath{\textasteriskcentered}{*}
  \fi
}

\begin{document}

\title{Cascade Prompt Learning for Vision-Language Model Adaptation} 

\titlerunning{Cascade Prompt Learning}


\author{Ge Wu\inst{1}$^\dag$\orcidlink{0009-0008-3011-091X},
Xin Zhang\inst{1}$^\dag$\orcidlink{0009-0000-9078-9110},
Zheng Li\inst{1}\orcidlink{0000-0003-3309-1087},
Zhaowei Chen\inst{3}\orcidlink{0000-0002-9508-6999},
\\Jiajun Liang\inst{3}\orcidlink{0000-0001-5586-340X},
Jian Yang\inst{1}\orcidlink{0000-0003-4800-832X},
Xiang Li\inst{1,2}$^*$\orcidlink{0000-0002-4996-7365}
}

\footnotetext[1]{Equal contributions. Work is done when Ge Wu is an intern at Megvii Technology.}%
\footnotetext[2]{Corresponding author.}%

\authorrunning{G. Wu et al.}

\institute{VCIP, CS, Nankai University 
\and NKIARI, Shenzhen Futian
\and Megvii Technology
\\ \email{gewu.nku@gmail.com, \{zhasion, zhengli97\}@mail.nankai.edu.cn, \\ \{csjyang, xiang.li.implus\}@nankai.edu.cn, \\ \{chenzhaowei, liangjiajun\}@megvii.com}
}

\maketitle

\begin{abstract}
Prompt learning has surfaced as an effective approach to enhance the performance of Vision-Language Models~(VLMs) like CLIP when applied to downstream tasks. However, current learnable prompt tokens are primarily used for the single \implus{phase} of adapting to tasks \implus{(i.e., adapting prompt)}, easily leading to overfitting risks. In this work, we propose a novel \textbf{Cas}cade \textbf{P}rompt \textbf{L}earning (\textbf{CasPL}) framework to enable prompt learning to serve both generic and specific expertise \implus{(i.e., boosting and adapting prompt)} simultaneously. Specifically, CasPL is a new learning paradigm comprising two distinct phases of learnable prompts: the first boosting prompt is crafted to extract domain-general knowledge from a senior larger CLIP teacher model by aligning their predicted logits using extensive unlabeled domain images. The second adapting prompt is then cascaded with the frozen first set to fine-tune the downstream tasks, following the approaches employed in prior research. \implus{In this manner, CasPL can effectively capture both domain-general and task-specific representations into explicitly different gradual groups of prompts, thus potentially alleviating overfitting issues in the target domain. It's worth noting that CasPL serves as a plug-and-play module that can seamlessly integrate into any existing prompt learning approach. CasPL achieves a significantly better balance between performance and inference speed, which is especially beneficial for deploying smaller VLM models in resource-constrained environments.} Compared to the previous state-of-the-art method PromptSRC, CasPL shows an average improvement of 1.85$\%$ for base classes, 3.44$\%$ for novel classes, and 2.72$\%$ for the harmonic mean over 11 image classification datasets. Code is publicly available at: \url{https://github.com/megvii-research/CasPL}.
  \keywords{Prompt learning \and multi-phase \and plug-and-play}
\end{abstract}

\begin{figure}[t]
    \centering
    \includegraphics[width=1\linewidth]{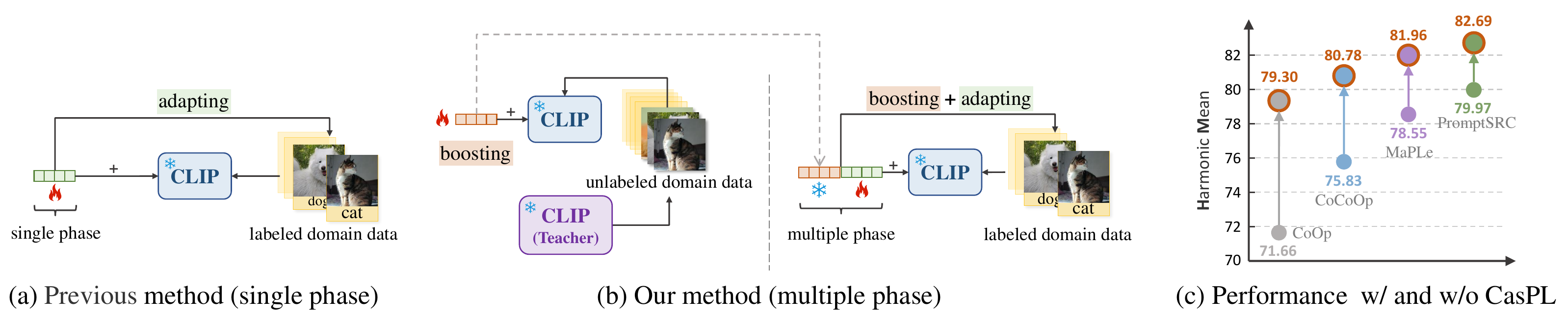}
    \caption{Comparison of CasPL with previous prompt learning methods. \indexbf{(a)} Previous methods adopt single phase prompting techniques for \hladapt{adapting} the domain datasets. \indexbf{(b)} CasPL introduces cascaded diverse prompts with multiple functions consisting of both \hlboost{boosting} and \hladapt{adapting} prompt phases. \indexbf{(c)} Performance (HM) of previous prompt learning methods \textit{w/} or \textit{w/o} our CasPL on base-to-novel tasks. The results are the average on 11 datasets.  
    }\label{fig:compare}
\end{figure}

\section{Introduction}
\label{sec:intro}
Vision-Language Models (VLMs) like CLIP~\cite{radford2021learning} have garnered significant attention recently due to their impressive generalization capabilities. Trained on an extensive dataset of image-text pairs using a contrastive loss, CLIP demonstrates robust representation skills under open vocabulary settings. Consequently, various studies~\cite{luddecke2022image,yu2023turning,zhong2022regionclip,gu2021open,ding2022decoupling} leverage pre-trained CLIP, fine-tuning it for domain-specific downstream tasks.
Among these fine-tuning methods, prompt learning~\cite{li2021prefix,brown2020language,lester2021power,liu2022p}
has gained prominence. This involves fixing the pre-trained model and adjusting only the input prompts. Initially employed in NLP as a text prompt to fine-tune large language models~\cite{devlin2018bert}, this approach has been validated and extended for applications in both vision~\cite{jia2022visual, wang2022dualprompt, wang2022learning} and vision-language~\cite{zhou2022learning,zhou2022conditional, khattak2023self,khattak2023maple,li2024promptkd} tasks.
A number of recent investigations~\cite{zhou2022learning,zhu2023prompt} have shown that utilizing adaptable continuous prompts consistently yields better outcomes than applying fixed text prompts.
\implus{Subsequently, the predominant research efforts~\cite{zang2022unified,zhou2022conditional,khattak2023maple, khattak2023self, lu2022prompt} in the field are primarily centered on developing dual vision-language or hierarchical prompt formulations to enhance the adaptability of CLIP models over downstream tasks, achieving impressive performance.}

\important{Despite their great success, it is noticeable that the adaptable prompt tokens in prior studies are predominantly employed for the singular phase of adapting to domain tasks (i.e., adapting prompt, see Fig.~\ref{fig:compare}(a)), thus easily leading to the overfitting problems.} Instead of the previous single-phase tuning paradigm of prompts, in this paper, we introduce a novel plug-and-play framework called \lz{\textbf{Cas}cade \textbf{P}rompt \textbf{L}earning (\textbf{CasPL})}, which incorporates two distinct sets of learnable prompts with multiple roles: boosting and adapting prompts. These prompts are optimized gradually across two phases. During the initial phase, the boosting prompts are learned to extract domain-general knowledge from a senior larger CLIP teacher model by aligning their prediction logits using extensive unlabeled domain image data (see Fig.~\ref{fig:compare}(b) \lz{left}). In the second phase, the adapting prompt is optimized by subsequently cascading with the fixed boosting prompt from the first phase to fine-tune the downstream tasks, following the approaches employed in prior research~\cite{zhou2022learning,zhu2023prompt,zang2022unified,zhou2022conditional,khattak2023maple, khattak2023self} (see Fig.~\ref{fig:compare}(b) \lz{right}).

CasPL has several unique advantages. \textbf{Firstly, the boosting prompt is optimized in an unsupervised manner, allowing it to harness a substantial amount of unlabeled domain data.} Specifically, the boosting prompt distills general, advanced knowledge from a senior larger CLIP teacher using unlabeled domain images. This knowledge inherently comprises general domain priors, which, in turn, fortify the original CLIP model against the risks of overfitting in this domain~(Fig.~\ref{fig:compare}(c)). \textbf{Secondly, CasPL is a plug-and-play framework that can be incorporated into any existing prompt learning methodologies.} \implus{The boosting prompt enhances the adaptability of the original CLIP model to the target domain data with very few parameters {($<0.1\%$)} and negligible inference cost. The frozen CLIP model with the fixed boosting prompts can be regarded as a new ``original'' CLIP model, analogous to the mathematical concept of ``change of variables'' (CLIP $\longleftarrow$ CLIP + boosting prompt). Therefore, the updated ``original'' CLIP model can naturally be adapted to any existing prompt learning methods.} \rewrite{\textbf{Thirdly, CasPL enables a smaller model (ViT-B/16) to match the performance of a larger model~(ViT-L/14) while maintaining efficient inference.} By incorporating the frozen boosting prompt into the smaller model and training the adapting prompt with any prompt learning method, CasPL achieves a better balance between inference speed and performance. This is particularly beneficial for deploying models in resource-constrained settings, where only smaller models are feasible.}

Our contributions can be summarized as follows:
\begin{itemize}[leftmargin=20pt]\setlength{\itemsep}{5pt}
    \item \implus{We propose a novel \lz{cascade prompt learning framework consisting of both boosting and adapting prompt phases. 
    To our best knowledge,}
    CasPL is the first to introduce cascaded diverse prompts with multiple phases for VLMs, which is a brand new learning paradigm for fine-tuning VLMs.}

    \item \implus{We demonstrate that the boosting prompts can distill domain-general knowledge from the senior teacher over massive unlabeled domain images, leading to superior recognition performance and \rewrite{efficient inference.}}

    \item \implus{As a plug-and-play framework, CasPL can be seamlessly integrated into any existing prompt learning approaches, with negligible parameters (boosting prompt tokens, {$<0.1\%$}) and ignorable additional inference cost introduced.}

    \item Compared to the previous state-of-the-art method PromptSRC, CasPL shows an average improvement of 1.85$\%$ for base classes, 3.44$\%$ for novel classes, and 2.72$\%$ for the harmonic mean over 11 image classification datasets. 
\end{itemize}

\section{Related Work}
\textbf{Vision Language Models.}
Foundational Vision-Language Models~(VLMs) like CLIP~\cite{radford2021learning} and ALIGN~\cite{jia2021scaling} have demonstrated significant advancements in recent years across a diverse spectrum of tasks~\cite{luddecke2022image, yu2023turning, zhong2022regionclip, gu2021open, ding2022decoupling, gao2023clip, zhang2021tip}. 
A representative work is CLIP, which employs a contrastive loss to simultaneously optimize two encoders, enabling the mapping of both images and text to a shared embedding space. This space facilitates the alignment of vision and language representations between paired images and text. 
Furthermore, the success of VLMs relies heavily on the availability of substantial training data—for instance, CLIP and ALIGN leverage 400 million and 1 billion network image-text pairs. Due to the extensive training data, pre-trained VLMs exhibit robust vision representation capabilities for open vocabulary. 
Consequently, zero-shot transfer learning becomes readily accessible for addressing various vision tasks.

\noindent{\textbf{Prompt Learning.}}
Prompt learning represents a novel training paradigm in the realm of NLP~\cite{liu2023pre}. This approach streamlines the training process by necessitating input adjustments rather than fine-tuning all parameters in a pre-trained model. Fine-tuning necessitates manual prompt design primitively~\cite{brown2020language, schick2020few, gao2020making}, fraught with challenges and instability when relying on natural language-based discrete prompts~\cite{zhao2021calibrate, lester2021power}. 
Recent developments bypass these discrete prompts instead of focusing on learning continuous prompts to replace their predecessors~\cite{li2021prefix}. Inspired by the success of prompt learning in NLP, researchers have also shown its applicability in vision tasks~\cite{jia2022visual, wang2022learning}.
To enhance the efficiency of solving downstream tasks using VLMs, CoOp~\cite{zhou2022learning} introduces learnable prompts within the language branch of CLIP for model fine-tuning. Recognizing the multimodal nature of VLMs, MaPLe~\cite{khattak2023maple}, and UPT~\cite{zang2022unified} employ multimodal information interactions for prompt learning. In addition, some work focuses on solving overfitting problems during fine-tuning. CoCoOp~\cite{zhou2022conditional} introduces a conditional prompt based on visual features. ProGrad~\cite{zhu2023prompt} proposes a prompt alignment gradient to prevent prompt tuning from forgetting the general knowledge. PromptSRC~\cite{khattak2023self} utilizes a regularization framework to ensure the model maintains generality while adapting to specific tasks. DePT~\cite{zhang2023dept} decouples base-specific knowledge from feature channels into an isolated feature space. \implus{It is crucial to highlight that all current methods focus on optimizing learnable prompts in a single phase. In contrast, our research takes a unique approach by initially examining various roles and phases of prompt tokens gradually, resulting in the development of a novel framework termed Cascade Prompt Learning~(CasPL).}

\noindent{\textbf{Knowledge Distillation in VLMs.}}
{The objective of knowledge distillation~\cite{hinton2015distilling,tian2019contrastive,jiao2019tinybert,li2021online,yang2022mutual,li2023curriculum,WenhuaZhang-MOT1} is to transfer the expertise of a teacher model to a student model, thereby enhancing the performance of the student model. Recently, there has been a surge of research investigating utilizing knowledge distillation with VLMs~\cite{fang2021compressing, wang2022multimodal, yang2023clip, wang2022clip, wu2023tinyclip, li2024promptkd}.
For instance, LP-CLIP~\cite{laroudie2023improving} incorporates a learnable linear probing layer for knowledge distillation, CLIP-KD~\cite{yang2023clip} explores the effects of various distillation strategies by pre-training the weights of the student CLIP model, and Tiny-CLIP~\cite{wu2023tinyclip} trains a minor student CLIP model via affinity mimicking and weight inheritance during the pre-training stage. \implus{In contrast to these previous approaches, the distillation in our first phase is designed for domain distillation instead of large-scale pre-training.} Furthermore, there have been studies focusing on distilling the capabilities of CLIP into traditional CNN~\cite{he2016deep,he2017mask}/ViT~\cite{dosovitskiy2020image} architectures for knowledge distillation in tasks such as open-vocabulary object detection~\cite{gu2021open,liu2023efficient,zhong2022regionclip} and semantic segmentation~\cite{jiao2023learning}, \implus{which differs from the CLIP-based teacher-student paradigm used in this work.}

\noindent{\textbf{Unsupervised Learning in VLMs.}}
There is a trend in incorporating unsupervised learning into prompt learning for VLMs by adopting the concept of pseudo-labels~\cite{menghini2023enhancing, huang2022unsupervised, laroudie2023improving, mirza2023lafter, li2024promptkd}. Pseudo-labels were initially developed as a semi-supervised technique~\cite{lee2013pseudo}, requiring a portion of labeled data to train a baseline model for generating these labels. However, the emergence of VLMs has rendered the need for this labeled data obsolete. UPL~\cite{huang2022unsupervised} and LaFTer~\cite{mirza2023lafter} employ unsupervised learning in prompt learning to generate pseudo-labels for target datasets. ENCLIP~\cite{menghini2023enhancing} conducts unsupervised training using iteratively refined pseudo-labels generated by CLIP.
Instead of leveraging the pseudo-labels, our work combines unsupervised learning and knowledge distillation in the prompt learning of VLMs, using the unlabeled domain data.

\begin{figure*}[t]
    \centering
    \includegraphics[width=1\linewidth]{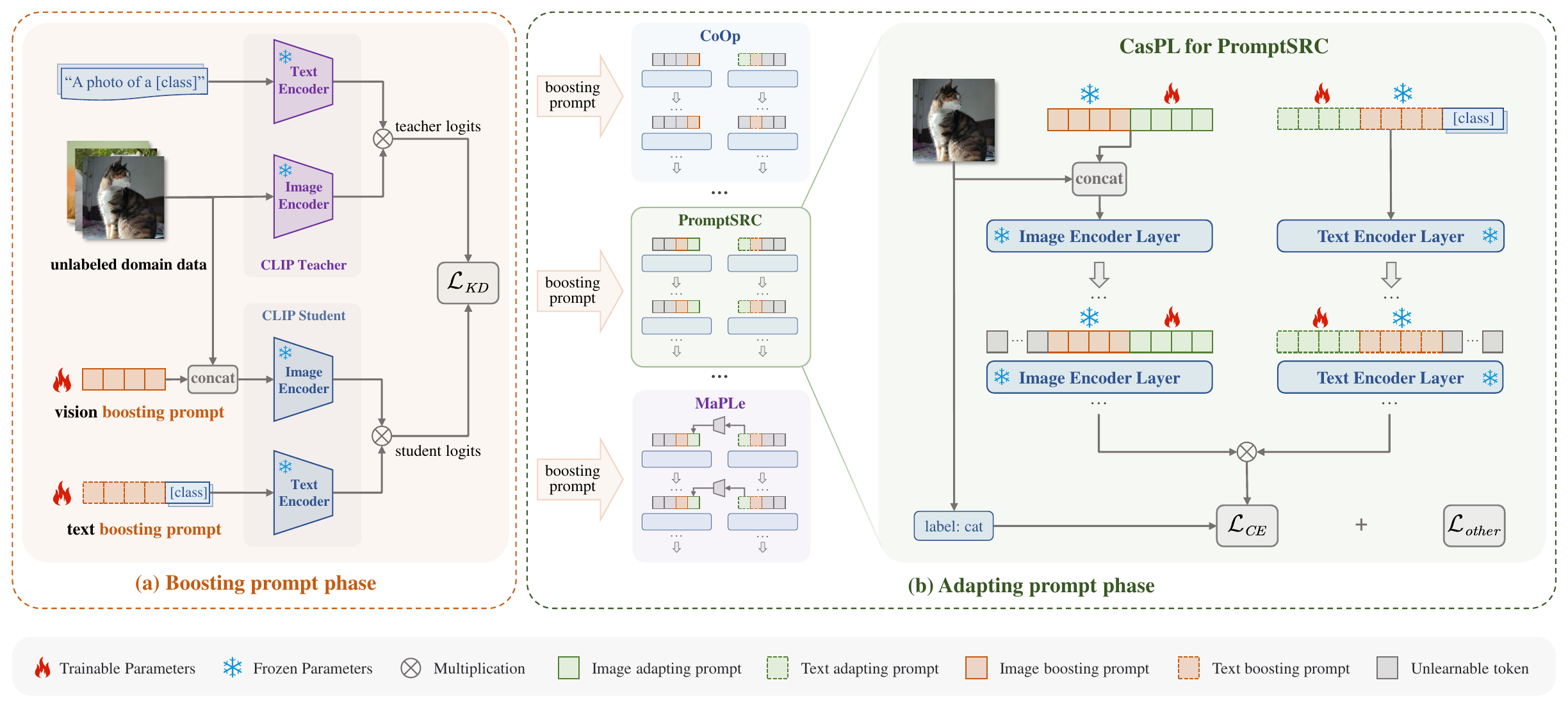}
    \caption{An overview of our proposed CasPL framework. \indexbf{(a)} We utilize a set of \hlboost{boosting} prompts to enable the student CLIP model to extract general domain knowledge from the teacher CLIP model, leveraging an extensive amount of unlabeled domain data. \indexbf{(b)} The boosting prompt can be seamlessly incorporated into existing related work as a plug-in. Here, we exemplify this integration with PromptSRC, where \emph{frozen} \hlboost{boosting} prompts are cascaded with \emph{learnable} \hladapt{adapting} prompts without altering any loss function. Further details regarding adaptations to other methods (e.g., CoOp~\cite{zhou2022learning}, CoCoOp~\cite{zhou2022conditional}, MaPLe~\cite{khattak2023maple}) are provided in the Appendix. 
    }
	\label{fig:framework}
\end{figure*}

\section{Method}
\label{sec:method}
\implus{We develop a method that investigates and validates the potential of prompts for diverse functions. 
An overview of our Cascade Prompt Learning~(CasPL) framework is presented in Fig.~\ref{fig:framework}. Unlike previous approaches, our method distinctly outlines multiple phases for prompts. In the following sections, we delve into the CLIP architecture and prompt learning method in Sec.\ref{sec:preliminaries}. Subsequently, we provide a detailed introduction to the proposed CasPL framework in Sec.~\ref{sec:CasPL}.}

\subsection{Preliminaries}
\label{sec:preliminaries}
Our approach is built upon the foundation of the pre-trained CLIP~\cite{radford2021learning}. Specifically, we employ vision transformer~(ViT)~\cite{dosovitskiy2020image}  based CLIP models, characterized by a text encoder and an image encoder. 

In the image encoder, an image $\boldsymbol{I}\in\mathbb{R}^{H\times W\times 3}$ is divided into patches and then projected into patch embeddings. These patch embeddings serve as the input to transformer blocks, yielding image features $\boldsymbol{x}\in\mathbb{R}^d$ from the final output. On the other hand, for the text encoder, the input $\boldsymbol{T}$ is typically a fixed template such as ${``a\ photo\ of\ a\ [class]"}$, where ${[class]}$ signifies the corresponding category. This input is tokenized into word embeddings, which are then input into transformer blocks, resulting in text features $\boldsymbol{y}\in\mathbb{R}^d$ from the final output. \implus{By computing the similarities between the image feature and a set of text features,} we can establish a zero-shot classifier via pre-trained CLIP, and the resulting prediction probability is:
\begin{equation}
q(c|\boldsymbol{x})=\frac{\exp{(\mathcal{S}({\boldsymbol{x}},\ {\boldsymbol{y_c}}) \ / \ \tau)}}{\sum_{i=1}^C\exp{(\mathcal{S}({\boldsymbol{x}},\ {\boldsymbol{y_i}}) \ / \ \tau)}}\ ,
\end{equation}
where $\mathcal{S}(\cdot,\cdot)$ represents cosine similarity, $c$ denotes a category and $\boldsymbol{y_c}$ denotes the corresponding text feature, $\boldsymbol{y_i}$ indicates the text feature of $i_{th}$ category, $C$ denotes the number of categories, and $\tau$ is a temperature hyper-parameter~\cite{hinton2015distilling}.

Instead of utilizing a manual text template, existing prompt learning methods~\cite{li2021prefix, zhou2022learning} append a set of learnable prompts into the input of the text encoder. The text template ``${a\ photo\ of\ a\ [class]}$'' is replaced by ``${\color{adapt_line}{{\mathrm{p}_1^t \ \mathrm{p}_2^t\dots \mathrm{p}_N^t}}} \ [class]$'', where ${\color{adapt_line}{\mathrm{p}^t_i}}$ $(i \in 1,\dots,N)$ denotes a learnable prompt for text branch, and $N$ denotes the number of learnable prompts. Similarly, we can append another set of learnable tokens after the patch embeddings, like $\{I_{cls}, I_1, I_2\dots I_M, {\color{adapt_line}{\mathrm{p}^v_1,\mathrm{p}_2^v\dots \mathrm{p}_N^v}} \}$, where $I_{cls}$ denotes a class token, $I_i(i\in1,\dots, M)$ denotes a patch embedding, $M$ denotes the number of patch embedding, ${\color{adapt_line}{\mathrm{p}^v_i}}\ (i\in1,\dots, N)$ denotes a learnable token for vision branch, and $N$ denotes the number of learnable prompts. Several studies~\cite{khattak2023maple,jia2022visual,khattak2023self} demonstrate the effectiveness of incorporating learnable prompts into various depths within both the image and text encoders. \important{Therefore, we follow these practices when designing the formats of boosting prompts.}

\subsection{Cascade Prompt Learning}
\label{sec:CasPL}
{To seek better generalization ability in each domain-specific downstream task, we propose a new learning paradigm that divides the training process into two phases to investigate the multiple roles of prompts for serving as generic and specific experts.
In the initial phase, the boosting prompts are employed to extract advanced domain-general knowledge from a larger CLIP teacher, utilizing unlabeled domain images. Following this, in the second phase, the adapting prompts are cascaded with the frozen first set of boosting prompts to address domain-specific downstream tasks effectively. As a result, CasPL enables a smaller model (ViT-B/16) to perform as well as a larger model (ViT-L/14).

\subsubsection{Boosting Prompt Phase}
We incorporate the larger CLIP teacher model to generate a set of boosting prompts for the target CLIP student model. As mentioned in Sec.~\ref{sec:preliminaries}, for the student CLIP model, we utilize ``${\color{boost_line}{p_1^t \ p_2^t\dots p_L^t}}\ [class]$'' as the input of its text encoder, and $\{I_{cls},I_1,I_2\dots I_M, {\color{boost_line}{p^v_1,p_2^v\dots p_L^v}} \}$ for its image encoder, where the \textbf{\textcolor{boost_line}{brown}} color denotes the learnable boosting prompts, and $L$ denotes the length of boosting prompt tokens. Our goal is to refine its understanding of domain-general knowledge by fine-tuning vision and textual boosting prompts using plentiful unlabeled domain data (see Fig.~\ref{fig:framework} left). The predicted logits of the student, denoted as $f^S$, are derived by multiplying the normalized features from both vision and \implus{a set of texts from the domain categories}. For the teacher model, we employ a straightforward text template ${``a\ photo\ of\ a\ [class]"}$ for the text encoder \implus{in most cases (more details refer to the Appendix)}. We represent the teacher's predicted logits as $f^T$. The loss function, designed to align their predicted logits through the utilization of massive unlabeled domain images, can be expressed as follows:
\begin{equation}
\label{eq_kd}
\mathcal{L}_{KD}(f^S, f^T)=KL(\sigma({f^T}/\tau), \sigma(f^S/\tau))\ .
\end{equation}
Here, $\sigma(\cdot)$ represents the softmax operation, $\tau$ is a temperature hyper-parameter, and $KL(\cdot,\cdot)$ refers to the Kullback-Leibler divergence loss. The entire boosting prompts, including text prompts $\{{\color{boost_line}{p_1^t \ p_2^t\dots p_L^t}}\}$ and vision prompts $\{{\color{boost_line}{p^v_1,p_2^v\dots p_L^v}}\}$, are optimized by Eq.~\eqref{eq_kd} during the first phase.

\subsubsection{Adapting Prompt Phase}
The proposed CasPL involves cascading the frozen boosting prompts with the adapting prompts. Specifically, the fixed boosting prompt learned from the initial phase is a plug-and-play prompt module that can seamlessly integrate into any existing prompt learning methods (see Fig.~\ref{fig:framework} middle). When it cooperates with previous prompt learning frameworks, the input to the text encoder is extended to ``${\color{adapt_line}{\mathrm{p}_1^t \ \mathrm{p}_2^t\dots \mathrm{p}_N^t}},{\color{boost_line}{p_{1}^t \dots p_{L}^t}} \ [class]$''. For the image encoder, it becomes $\{I_{cls},I_1,I_2\dots I_M, {\color{boost_line}{p^v_1,p_2^v\dots p_L^v}},{\color{adapt_line}{\mathrm{p}^v_{1},\dots \mathrm{p}_{N}^v}} \}$. Here, the \textbf{\textcolor{boost_line}{brown}} color represents the \emph{frozen} boosting prompts from the first phase, while the \textbf{\textcolor{adapt_line}{green}} color represents the \emph{learnable} adapting prompts in this phase. 
In this phase, we aim to learn the adapting prompt through the supervision of labeled (few-shot) images as before, without altering any loss function \implus{of existing prompt learning frameworks} (see Fig.~\ref{fig:framework} right).

\begin{table*}[!t]
    \centering
    \caption{Comparison with state-of-the-art methods \textit{w/} or \textit{w/o} CasPL on base-to-novel generalization. CasPL consistently improves model performance on 11 datasets.}
    \begin{subtable}[t]{0.32\linewidth}
        \resizebox{0.96\linewidth}{!}
        {
        \begin{tabular}{c|ccl}
            \hline
            Method    & Base  & Novel & HM \\
            \hline
            CLIP & 69.34 & 74.22 & 71.70 \\
            \hline
            CoOp & 82.69 & 63.22 & 71.66 \\
            \rowcolor{tablecolor} \textbf{+CasPL} &84.78 	&74.49 	&79.30    \improve{(+7.64)} \\
            \hline
            CoCoOp & 80.47 & 71.69 & 75.83 \\
            \rowcolor{tablecolor} \textbf{+CasPL} &83.63 	&78.12 	&80.78   \improve{(+4.95)} \\
            \hline
            MaPLe   & 82.28 & 75.14 & 78.55 \\
            \rowcolor{tablecolor} \textbf{+CasPL} &84.48 &79.59 	&81.96  \improve{(+3.41)}     \\
            \hline
            PromptSRC & 84.26 & 76.10 & 79.97 \\
            \rowcolor{tablecolor} \textbf{+CasPL} &86.11  &79.54 &82.69  \improve{(+2.72)}   \\
            \hline
        \end{tabular}
        }
        \caption{Average over 11 datasets.}
    \end{subtable}
    \begin{subtable}[t]{0.32\linewidth}
        \resizebox{0.96\linewidth}{!}
        {
        \begin{tabular}{c|ccl}
            \hline
            Method    & Base  & Novel & HM \\
            \hline
            CLIP & 72.43 & 68.14 & 70.22 \\
            \hline
            CoOp & 76.47 & 67.88 & 71.92 \\
            \rowcolor{tablecolor} \textbf{+CasPL} &77.90 	&67.43 	&72.29    \improve{(+0.37)}  \\
            \hline
            CoCoOp & 75.98 & 70.43 & 73.10 \\
            \rowcolor{tablecolor} \textbf{+CasPL} &77.40 	&71.40 	&74.28   \improve{(+1.18)} \\
            \hline
            MaPLe   & 76.66 & 70.54 & 73.47 \\
            \rowcolor{tablecolor} \textbf{+CasPL} &78.20 	&71.47 	&74.68  \improve{(+1.21)}\\
            \hline
            PromptSRC & 77.60 & 70.73 & 74.01 \\
            \rowcolor{tablecolor} \textbf{+CasPL} &78.97 	&70.50 	&74.50  \improve{(+0.49)} \\
            \hline
        \end{tabular}
        }
        \caption{ImageNet}
    \end{subtable}
    \begin{subtable}[t]{0.32\linewidth}
        \resizebox{0.96\linewidth}{!}
        {
        \begin{tabular}{c|ccl}
            \hline
            Method    & Base  & Novel & HM \\
            \hline
            CLIP & 96.84 & 94.00 & 95.40 \\
            \hline
            CoOp & 98.00 & 89.81 & 93.73 \\
            \rowcolor{tablecolor} \textbf{+CasPL} &98.63 	&95.50 	&97.04     \improve{(+3.31)}  \\
            \hline
            CoCoOp & 97.96 & 93.81 & 95.84 \\
            \rowcolor{tablecolor} \textbf{+CasPL} &98.60 	 &94.63 	 &96.58    \improve{(+0.74)} \\
            \hline
            MaPLe & 97.74 & 94.36 & 96.02 \\
            \rowcolor{tablecolor} \textbf{+CasPL} &98.47 	&96.03 	&97.23   \improve{(+1.21)} \\
            \hline
            PromptSRC & 98.10 & 94.03 & 96.02 \\
            \rowcolor{tablecolor} \textbf{+CasPL} &98.60 	&95.70 	&97.13   \improve{(+1.11)} \\
            \hline
        \end{tabular}
        }
        \caption{Caltech101}
    \end{subtable}
    
    
    \begin{subtable}[t]{0.32\linewidth}
        \resizebox{0.96\linewidth}{!}
        {
        \begin{tabular}{c|ccl}
            \hline
            Method    & Base  & Novel & HM \\
            \hline
            CLIP & 91.17 & 97.26 & 94.12 \\
            \hline
            CoOp & 93.67 & 95.29 & 94.47 \\
            \rowcolor{tablecolor} \textbf{+CasPL} &94.83 	&97.37 	&96.08    \improve{(+1.61)}  \\
            \hline
            CoCoOp & 95.20 & 97.69 & 96.43 \\
            \rowcolor{tablecolor} \textbf{+CasPL} &95.23 	&98.10 	&96.65   \improve{(+0.22)}  \\
            \hline
            MaPLe & 95.43 & 97.76 & 96.58 \\
            \rowcolor{tablecolor} \textbf{+CasPL} &95.37 	&98.13 	&96.73  \improve{(+0.15)}  \\
            \hline
            PromptSRC & 95.33 & 97.30 & 96.30 \\
            \rowcolor{tablecolor} \textbf{+CasPL} &95.53 	&98.07 	&96.78  \improve{(+0.48)}  \\
            \hline
        \end{tabular}
        }
        \caption{OxfordPets}
    \end{subtable}
    \begin{subtable}[t]{0.32\linewidth}
        \resizebox{0.96\linewidth}{!}
        {
        \begin{tabular}{c|ccl}
            \hline
            Method    & Base  & Novel & HM \\
            \hline
            CLIP & 63.37 & 74.89 & 68.65 \\
            \hline
            CoOp & 78.12 & 60.40 & 68.13 \\
            \rowcolor{tablecolor} \textbf{+CasPL} &80.23 	&75.77 	&77.94    \improve{(+9.81)}  \\
            \hline
            CoCoOp & 70.49 & 73.59 & 72.01 \\
            \rowcolor{tablecolor} \textbf{+CasPL} &76.37 	&81.07 	&78.65   \improve{(+6.64)}  \\
            \hline
            MaPLe & 72.94 & 74.00 & 73.47 \\
            \rowcolor{tablecolor} \textbf{+CasPL} &78.10 	&81.97 	&79.99  \improve{(+6.52)} \\
            \hline
            PromptSRC & 78.27 & 74.97 & 76.58 \\
            \rowcolor{tablecolor} \textbf{+CasPL} &82.33 	&79.93 	&81.11  \improve{(+4.53)} \\
            \hline
        \end{tabular}
        }
        \caption{StanfordCars}
    \end{subtable}
    \begin{subtable}[t]{0.32\linewidth}
        \resizebox{0.96\linewidth}{!}
        {
        \begin{tabular}{c|ccl}
            \hline
            Method    & Base  & Novel & HM \\
            \hline
            CLIP & 72.08 & 77.80 & 74.83 \\
            \hline
            CoOp & 97.60 & 59.67 & 74.06 \\
            \rowcolor{tablecolor} \textbf{+CasPL} &98.27 	&73.47 	&84.08    \small{\improve{(+10.02)}}   \\
            \hline
            CoCoOp & 94.87 & 71.75 & 81.71 \\
            \rowcolor{tablecolor} \textbf{+CasPL} &96.63 	&76.50 	&85.40   \improve{(+3.69)}  \\
            \hline
            MaPLe & 95.92 & 72.46 & 82.56 \\
            \rowcolor{tablecolor} \textbf{+CasPL} &97.73 	&77.63 	&86.53  \improve{(+3.97)}   \\
            \hline
            PromptSRC & 98.07 & 76.50 & 85.95 \\
            \rowcolor{tablecolor} \textbf{+CasPL} &98.73 	&80.13 	&88.46  \improve{(+2.51)} \\
            \hline
        \end{tabular}
        }
        \caption{Flowers102}
    \end{subtable}


    \begin{subtable}[t]{0.32\linewidth}
        \resizebox{0.96\linewidth}{!}
        {
        \begin{tabular}{c|ccl}
            \hline
            Method    & Base  & Novel & HM \\
            \hline
            CLIP & 90.10 & 91.22 & 90.66 \\
            \hline
            CoOp & 88.33 & 82.26 & 85.19 \\
            \rowcolor{tablecolor} \textbf{+CasPL} &90.60 	&91.03 	&90.82    \improve{(+5.63)}   \\
            \hline
            CoCoOp & 90.70 & 91.29 & 90.99 \\
            \rowcolor{tablecolor} \textbf{+CasPL} &91.50 	&92.93 	&92.21   \improve{(+1.22)}  \\
            \hline
            MaPLe & 90.71 & 92.05 & 91.38 \\
            \rowcolor{tablecolor} \textbf{+CasPL} &91.43 	&92.93 	&92.18  \improve{(+0.80)}  \\
            \hline
            PromptSRC & 90.67 & 91.53 & 91.10 \\
            \rowcolor{tablecolor} \textbf{+CasPL} &91.27 	&92.20 	&91.73  \improve{(+0.63)}  \\
            \hline
        \end{tabular}
        }
        \caption{Food101}
    \end{subtable}
    \begin{subtable}[t]{0.32\linewidth}
        \resizebox{0.96\linewidth}{!}
        {
        \begin{tabular}{c|ccl}
            \hline
            Method    & Base  & Novel & HM \\
            \hline
            CLIP & 27.19 & 36.29 & 31.09 \\
            \hline
            CoOp & 40.44 & 22.30 & 28.75 \\
            \rowcolor{tablecolor} \textbf{+CasPL} &45.83 	&37.30 	&41.13     \small\improve{(+12.38)}  \\
            \hline
            CoCoOp & 33.41 & 23.71 & 27.74 \\
            \rowcolor{tablecolor} \textbf{+CasPL} &42.93 	&41.23 	&42.07    \small\improve{(+14.33)} \\
            \hline
            MaPLe & 37.44 & 35.61 & 36.50 \\
            \rowcolor{tablecolor} \textbf{+CasPL} &43.60 	&42.20 	&42.89   \improve{(+6.39)}  \\
            \hline
            PromptSRC & 42.73 & 37.87 & 40.15 \\
            \rowcolor{tablecolor} \textbf{+CasPL} &48.23 	&41.97 	&44.88   \improve{(+4.73)} \\
            \hline
        \end{tabular}
        }
        \caption{FGVCAircraft}
    \end{subtable}
    \begin{subtable}[t]{0.32\linewidth}
        \resizebox{0.96\linewidth}{!}
        {
        \begin{tabular}{c|ccl}
            \hline
            Method    & Base  & Novel & HM \\
            \hline
            CLIP & 69.36 & 75.35 & 72.23 \\
            \hline
            CoOp & 80.60 & 65.89 & 72.51 \\
            \rowcolor{tablecolor} \textbf{+CasPL} &81.77 	&72.77 	&77.00    \improve{(+4.49)}   \\
            \hline
            CoCoOp & 79.74 & 76.86 & 78.27 \\
            \rowcolor{tablecolor} \textbf{+CasPL} &80.20 	&79.10 	&79.65   \improve{(+1.38)}  \\
            \hline
            MaPLe & 80.82 & 78.70 & 79.75 \\
            \rowcolor{tablecolor} \textbf{+CasPL} &82.23 	&79.80 	&81.00  \improve{(+1.25)} \\
            \hline
            PromptSRC & 82.67 & 78.47 & 80.52 \\
            \rowcolor{tablecolor} \textbf{+CasPL} &83.10 	&79.53 	&81.28  \improve{(+0.76)}  \\
            \hline
        \end{tabular}
        }
        \caption{SUN397}
    \end{subtable}


    \begin{subtable}[t]{0.32\linewidth}
        \resizebox{0.96\linewidth}{!}
        {
        \begin{tabular}{c|ccl}
            \hline
            Method    & Base  & Novel & HM \\
            \hline
            CLIP & 53.24 & 59.90 & 56.37 \\
            \hline
            CoOp & 79.44 & 41.18 & 54.24 \\ 
            \rowcolor{tablecolor} \textbf{+CasPL} &82.57 	&54.23 	&65.47     \small\improve{(+11.23)}  \\
            \hline
            CoCoOp & 77.01 & 56.00 & 64.85 \\
            \rowcolor{tablecolor} \textbf{+CasPL} &80.57 	&62.43 	&70.35    \improve{(+5.50)}  \\
            \hline
            MaPLe & 80.36 & 59.18 & 68.16 \\
            \rowcolor{tablecolor} \textbf{+CasPL} &82.73 	&66.77 	&73.90   \improve{(+5.74)} \\
            \hline
            PromptSRC & 83.37 & 62.97 & 71.75 \\
            \rowcolor{tablecolor} \textbf{+CasPL} &84.73     &69.63  &76.44   \improve{(+4.69)} \\
            \hline
        \end{tabular}
        }
        \caption{DTD}
    \end{subtable}
    \begin{subtable}[t]{0.32\linewidth}
        \resizebox{0.96\linewidth}{!}
        {
        \begin{tabular}{c|ccl}
            \hline
            Method    & Base  & Novel & HM \\
            \hline
            CLIP & 56.48 & 64.05 & 60.03 \\
            \hline
            CoOp & 92.19 & 54.74 & 68.69 \\
            \rowcolor{tablecolor} \textbf{+CasPL} &94.80 	&82.23 	&88.07     \small\improve{(+19.38)}  \\
            \hline
            CoCoOp & 87.49 & 60.04 & 71.21 \\
            \rowcolor{tablecolor} \textbf{+CasPL} &94.50 	 &83.63 	 &88.74    \small\improve{(+17.53)} \\
            \hline
            MaPLe & 94.07 & 73.23 & 82.35 \\
            \rowcolor{tablecolor} \textbf{+CasPL} &94.60 	&89.40 	&91.93   \improve{(+9.58)} \\
            \hline
            PromptSRC & 92.90 & 73.90 & 82.32 \\
            \rowcolor{tablecolor} \textbf{+CasPL} &96.67 	&85.87 	&90.95   \improve{(+8.63)} \\
            \hline
        \end{tabular}
        }
        \caption{EuroSAT}
    \end{subtable}
    \begin{subtable}[t]{0.32\linewidth}
        \resizebox{0.96\linewidth}{!}
        {
        \begin{tabular}{c|ccl}
            \hline
            Method    &  Base  & Novel & HM \\
            \hline
            CLIP & 70.53 & 77.50 & 73.85 \\
            \hline
            CoOp & 84.69 & 56.05 & 67.46 \\
            \rowcolor{tablecolor} \textbf{+CasPL} &87.10 	&72.30 	&79.01     \small\improve{(+11.55)}  \\
            \hline
            CoCoOp & 82.33 & 73.45 & 77.64 \\
            \rowcolor{tablecolor} \textbf{+CasPL} &86.00 	&78.33 	&81.99    \improve{(+4.35)} \\
            \hline
            MaPLe & 83.00 & 78.66 & 80.77 \\
            \rowcolor{tablecolor} \textbf{+CasPL} &86.83 	&79.10 	&82.79   \improve{(+2.02)} \\
            \hline
            PromptSRC & 87.10 & 78.80 & 82.74 \\
            \rowcolor{tablecolor} \textbf{+CasPL} &89.00 	&81.37 	&85.01   \improve{(+2.27)} \\
            \hline
        \end{tabular}
        }
        \caption{UCF101}
    \end{subtable}
    \label{table:base-to-novel}
\end{table*}

\begin{table}[t]
    \centering
    \caption{Domain generalization. The accuracies of CasPL on the source ImageNet dataset consistently demonstrate improvement. In other transferred domains, CasPL achieves remarkable enhancement on \rewrite{most ImageNet-V2 and ImageNet-S/-R.}}
    \resizebox{0.85\linewidth}{!}
    {
        \begin{tabular}{clllll}
        \hline
        \multirow{3}{*}{Method} & Source & \multicolumn{4}{c}{Target} \\
        \cmidrule(lr){2-2} \cmidrule(lr){3-6}
        & ImageNet & ImageNet-V2 & ImageNet-S  & ImageNet-R  & Average \\
        \hline
        CLIP & 66.73 & 60.83 & 46.15  & 73.96  &60.31  \\
        \hline
        CoOp & 71.51 & 64.20 & 47.99  & 75.21  &62.47  \\
        \rowcolor{tablecolor}\textbf{+CasPL} &71.91   \improve{(+0.40)}   &64.30   \improve{(+0.10)}   &48.29   \improve{(+0.30)}  &76.01   \improve{(+0.80)}    &62.87 \improve{(+0.40)} \\
        \hline
        CoCoOp & 71.02  & 64.07 & 48.75  & 76.18  &63.00  \\
        \rowcolor{tablecolor}\textbf{+CasPL} &71.31  \improve{(+0.28)}   &64.52  \improve{(+0.45)}  &48.20  \decline{(-0.55)}   &76.80  \improve{(+0.62)}  &63.17 \improve{(+0.17)} \\
        \hline
        MaPLe & 70.72 & 64.07 & 49.15  & 76.98  &63.40  \\
        \rowcolor{tablecolor}\textbf{+CasPL} &71.30 \improve{(+0.58)}    &64.29 \improve{(+0.22)}  & 48.81 \decline{(-0.34)}   &77.47 \improve{(+0.49)}    &63.52 \improve{(+0.12)}   \\
        \hline
        PromptSRC & 71.27 & 64.35 & 49.55  & 77.80   &63.90  \\
        \rowcolor{tablecolor}\textbf{+CasPL} &72.80 \improve{(+1.53)} &65.70 \improve{(+1.35)} & 49.71 \improve{(+0.16)} &77.90 \improve{(+0.10)}  &64.44 \improve{(+0.54)}\\
        \hline
        \end{tabular}
    }
    \label{table:domain-dataset}
\end{table}

\begin{table}[t]
    \centering
    \caption{\rewrite{Results of distinct unlabeled data source on 10 datasets. ``Out-domain'' means utilizing ImageNet as unlabeled out-domain data to fine-tune, while ``in-domain'' refers to using the corresponding dataset. CasPL improves few-shot image recognition with out-domain and in-domain data, highlighting its cross-domain solid generalization.}}
    \label{table:diff-domain-dataset}
    \setlength{\tabcolsep}{10pt}
    \resizebox{0.85\linewidth}{!}
    {
        \begin{tabular}{l|lll}
                \hline
                & PromptSRC                 &  \textbf{+CasPL (out-domain)} &  \textbf{+CasPL (in-domain)} \\
                \hline
                Acc. & \multicolumn{1}{c}{83.84} & \multicolumn{1}{c}{84.56 \improve{(+0.72\%)}}   & \multicolumn{1}{c}{85.52 \improve{(+1.68\%)}} \\
                \hline
        \end{tabular}
    }
\end{table}

\section{Experiments}
\noindent{\textbf{Datasets.}}
For base-to-novel generalization and few-shot experiments, we use 11 datasets following~\cite{zhou2022learning, zhou2022conditional}. Specifically, the datasets include ImageNet~\cite{deng2009imagenet} and Caltech101~\cite{fei2004learning} for generic objecting, FGVCAircraft~\cite{maji2013fine}, OxfordPets, StanfordCars~\cite{krause20133d}, Flowers102~\cite{nilsback2008automated}, and Food101~\cite{bossard2014food} for fine-grained classification, SUN397~\cite{xiao2010sun} for scene recognition,  EuroSAT~\cite{helber2019eurosat} satellite images classification, DTD~\cite{cimpoi2014describing} for describable texture classification, and UCF101~\cite{soomro2012ucf101} for action recognition. For domain generalization, we use ImageNet as the source dataset, and ImageNet-Sketch~\cite{wang2019learning}, ImageNet-V2~\cite{recht2019imagenet}, ImageNet-R~\cite{hendrycks2021natural} as the target dataset.

\noindent{\textbf{Training Details.}}
To align the comparisons with previous approaches~\cite{zhou2022learning,zhou2022conditional,khattak2023maple,khattak2023self}, we conduct experiments on ViT-B/16 CLIP model released by OpenAI~\cite{radford2021learning}. The vanilla ViT-L/14 CLIP model is adopted as the teacher model, and each dataset's entire training set (without class labels) is utilized as the unlabeled images in the first phase of CasPL. Following~\cite{zhou2022conditional,zhou2022learning}, the reported results are averaged over 3 runs. 
\important{Due to space limitations, more details of the experimental settings of CasPL are listed in the Appendix. }

\noindent{\textbf{Baselines.}}
We introduce our CasPL in a series of baseline methods, which consist of the single-modal prompt learning methods CoOp~\cite{zhou2022learning}, CoCoOp~\cite{zhou2022conditional}, and the multi-modal prompt learning methods PromptSRC~\cite{khattak2023self} and MaPLe~\cite{khattak2023maple}. Previously, PromptSRC obtained state-of-the-art performance by using its regularization framework. In addition, \important{for more fair comparisons,} we compare several CLIP adapting methods that utilize unlabeled domain data. These include three methods that only use unlabeled domain data to generate pseudo-label for fine-tuning, CLIP-PR~\cite{kahana2022improving}, UPL~\cite{huang2022unsupervised}, and LaFTer~\cite{mirza2023lafter}; three training strategies~\cite{menghini2023enhancing} implemented based on PromptSRC~\cite{khattak2023self}, which use few-shot base class data and unlabeled novel class data, FPL, IFPL and GRIP.

\begin{table}[t]
    \centering
    \caption{Comparisons with existing CLIP (ViT-B/16) adaptation methods which utilize unlabeled data, are evaluated on \rewrite{8 datasets}. These methods using few-shot labels (\cite{menghini2023enhancing}, ours) are all implemented based on PromptSRC~\cite{khattak2023self} for fair comparisons.  
    }
    \setlength{\tabcolsep}{10pt}
    \resizebox{0.7\linewidth}{!}
    {
        \begin{tabular}{l|c|ccc}
        \hline
        Method                                 & Publication    & Base   & Novel   & HM \\
        \hline
        CLIP~\cite{radford2021learning}        & ICML21          & 66.36  & 72.71   & 69.39   \\
        \hline
        CLIP-PR~\cite{kahana2022improving}     & ArXiv22         & 60.98  & 68.75   & 64.63 \\
        UPL~\cite{huang2022unsupervised}       & ArXiv22         & 68.82  & 75.82   & 72.15   \\
        LaFTer~\cite{mirza2023lafter}          & NeurIPS23       & 69.27  & 76.76   & 72.82   \\
        \hline
        FPL~\cite{menghini2023enhancing}       & NeurIPS23       & 83.63  &  74.87  & 79.01\\
        IFPL~\cite{menghini2023enhancing}      & NeurIPS23       & 84.08  &  76.08  & 79.88  \\
        GRIP~\cite{menghini2023enhancing}      & NeurIPS23       & 84.26  &  75.53  & 79.65  \\ 
        \textbf{CasPL (ours)  }                & --               &  \cellcolor{tablecolor}{\textbf{86.73}}  &  \cellcolor{tablecolor}\textbf{79.08}  & \cellcolor{tablecolor}\textbf{82.73}  \\ 
        \hline
        \end{tabular}
    }
    \label{table:weakly}
\end{table}

\subsection{Base-to-Novel Generalization}
We evaluate CasPL's generalizability by partitioning datasets into base and novel classes. The model is trained exclusively on the base classes in a few-shot setting and evaluated on both the base and novel categories.
Table~\ref{table:base-to-novel} shows the performance of the zero-shot CLIP~\cite{radford2021learning} and the previous methods \textit{w/} or \textit{w/o} our CasPL on 11 recognition datasets. These methods include CoOp~\cite{zhou2022learning}, CoCoOp~\cite{zhou2022conditional},  MaPLe~\cite{khattak2023maple}, and PromptSRC~\cite{khattak2023self}. As we can see, our CasPL consistently improves baseline model performance on various datasets. Compared to the previous state-of-the-art method, PromptSRC, CasPL demonstrates a 1.85$\%$ improvement in base classes, 3.44$\%$ improvement in novel classes, and 2.72$\%$ in harmonic mean. On average, the generalization improvement of the prior methods is primarily reflected in the enhancement of novel category accuracy. CasPL exhibits clear advantages in novel category accuracy improvement, with the 11.27$\%$ increase on CoOp, 6.43$\%$ on CoCoOp, and 4.45$\%$ on MaPLe. Notably, introducing the boosting prompt obtained in the first stage into CoOp as a plug-in achieves the same performance of vanilla PromptSRC.

\subsection{Domain Generalization}
\rewrite{Table~\ref{table:domain-dataset} compares results of previous methods \textit{with} and \textit{without} CasPL on cross-domain datasets, showing consistent enhancements by CasPL on the source datasets (0.40\% to 1.53\%). Additionally, CasPL demonstrates significant average improvements ranging from 0.12\% to 0.54\% on the target datasets. To further explore CasPL's domain generalization, Table~\ref{table:diff-domain-dataset} illustrates the performance across 10 datasets with various unlabeled data sources. CasPL (out-domain) utilizes ImageNet as unlabeled out-domain data, while CasPL (in-domain) uses the corresponding dataset. Using unlabeled out-domain and in-domain data results in 0.72\% and 1.68\% higher few-shot accuracy than PromptSRC, highlighting the efficacy of the two-stage decoupled training for cross-domain solid generalization.}

\subsection{Compare with un-/weakly-supervised methods}
\implus{Since CasPL involves the unlabeled domain data in the boosting prompt phase, we further conduct comprehensive experiments against the latest un-/weakly-supervised methods for CLIP.} \implus{These approaches~\cite{huang2022unsupervised,mirza2023lafter,menghini2023enhancing} typically adopt pseudo label techniques when leveraging the unlabeled data, which are quite different from CasPL.} All experiments utilize ViT-B/16 CLIP, and the default few-shot number is set to 16. Table~\ref{table:weakly} shows the performance of base-to-novel tasks on 8 datasets (except ImageNet, SUN397 and Food101) between CasPL and other compared methods. \rewrite{To ensure a fair comparison, we implement three pseudo-label training strategies (i.e., FPL, IFPL, and GRIP) in ENCLIP~\cite{menghini2023enhancing} based on the SOTA PromptSRC~\cite{khattak2023self} framework. By employing the same few-shot labeled data and unlabeled data, our method outperforms the best IFPL~\cite{menghini2023enhancing} strategy with a 2.85\% improvement in HM. Compared to methods~\cite{kahana2022improving, huang2022unsupervised, mirza2023lafter} using all unlabeled data, CasPL also demonstrates notable enhancements in HM (+9.90\%$\sim$18.10\%).} In general, approaches that leverage a combination of few-shot labeled domain data and a substantial amount of unlabeled domain data achieve better results than zero-shot strategies and methods that rely solely on unlabeled domain data.

\begin{figure*}[t]
	\centering
	\includegraphics[width=1\linewidth]{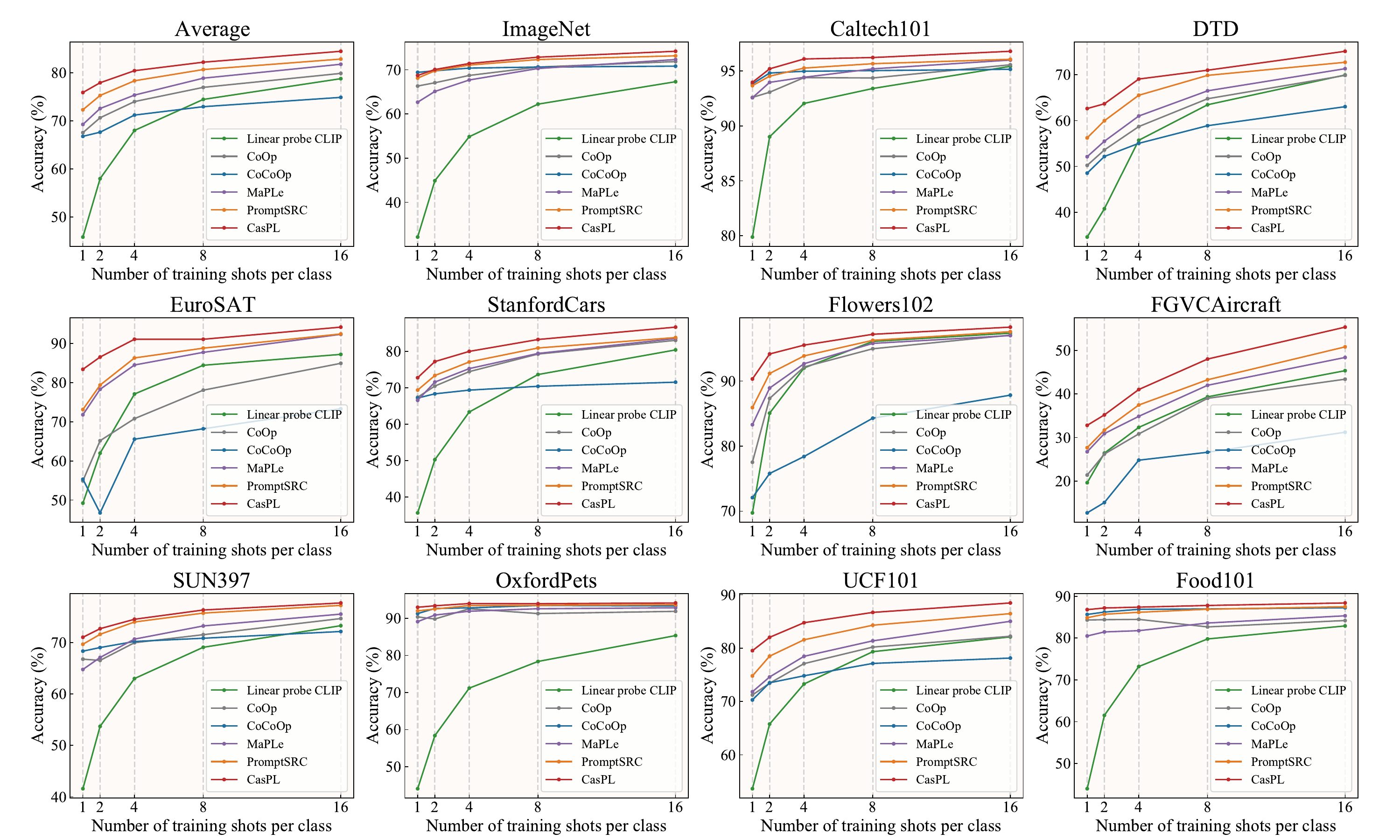}
	\caption{CasPL performance comparison in a few-shot image recognition setting. Based on PromptSRC, CasPL achieves the highest performance improvement across all settings.  These results emphasize the role of the initial boosting prompt of CasPL in extracting domain generalization capabilities from the senior larger CLIP.}
	\label{fig:fewshot}
\end{figure*}

\subsection{Few-shot Experiments}
We leverage the limited supervised image data to investigate whether the model can demonstrate good generalization across different $K$-shots per category, where $K$=1, 2, 4, 8, 16. The evaluation is conducted on the previous approaches and our CasPL based on PromptSRC, shown in Fig.~\ref{fig:fewshot}.
In terms of average results, our method consistently outperforms previous results across each shot setting, demonstrating the robustness of our approach.  Compared to the previous state-of-the-art method, PromptSRC, CasPL achieves performance gains of 3.59$\%$, 2.65$\%$, 2.1$\%$, 1.53$\%$, and 1.62$\%$ on 1, 2, 4, 8, and 16 shots across 11 datasets. Further, CasPL exhibits more pronounced advantages in extreme conditions ($K$=1). We attribute this to the boosting prompt enhancing domain generalization while the adapting prompt adeptly tailors the model to specific tasks.

\begin{table}[t]
    \centering
    \caption{Ablation study on layer depths in different phases using the harmonic mean metric. \important{A deeper layer consistently contributes to improved performance in general.}}
    \setlength{\tabcolsep}{10pt}
    \resizebox{0.8\linewidth}{!}
    {
        \begin{tabular}{cccccc}
        \hline
        \multirow{2}{*}{Layer depths of phase \text{I}} & \multicolumn{5}{c}{Layer depths of phase \text{II}}               \\ \cline{2-6} 
                                                 & 1 & 1$\sim$3 & 1$\sim$6 & 1$\sim$9 & 1$\sim$12 \\ \hline
        1                                        &79.95    & 80.79    &81.05   &81.24   &81.60            \\
        1$\sim$3                                 &80.78    &80.49     &81.69   &82.23   &82.25            \\
        1$\sim$6                                 &81.44    &81.92     &82.25      &82.18   &82.58           \\
        1$\sim$9                                 &82.11    &82.50      &82.76     &83.00   &83.19            \\
        1$\sim$12                                &82.30    &82.43      &82.83    &83.25    &\textbf{83.51}           \\
        \hline
        \end{tabular}
    }
    \label{table:ablation-depth}
\end{table}

\subsection{Ablation Studies}
\noindent{\textbf{Layer depth:}}
Table~\ref{table:ablation-depth} outlines the impact of employing different depths of learnable prompts in the two phases on performance. Our CasPL model, based on PromptSRC, is evaluated on 10 datasets (excluding the large-scale ImageNet) for experimental efficiency. The harmonic mean serves as the performance metric. 
Keeping the depth of the first phase constant while increasing the depth of the second phase results in an average performance improvement of approximately 1.94\% (from depth 1 to 12). Similarly, maintaining the depth of the second phase constant and augmenting the depth of the first phase leads to an average performance improvement of about 1.31\% (from depth 1 to 12). Overall, using learnable prompts with deeper layers in either phase enhances performance, aligning with prior research findings~\cite{jia2022visual, khattak2023maple}. The most optimal outcome is achieved when both phases employ a depth of 12 layers.

\noindent{\textbf{Prompt learnability and length:}}
Fig.~\ref{fig:ablation-length} illustrates the results addressing two inquiries: the optimal prompt length for CasPL and the necessity of freezing the boosting prompts in the second phase. Our observation indicates that when the prompt length remains constant, freezing the boosting prompts in the second phase always leads to superior performance compared to not freezing them. The primary objective of the first stage is to imbue prompts with generalization capabilities. Fine-tuning the boosting prompts to suit downstream tasks may compromise their inherent generalization ability within the domain. After establishing the learnability of boosting prompts, we determine that the optimal prompt length for both phases is 8.

\begin{figure}[t]
	\centering
	\includegraphics[width=.5\linewidth]{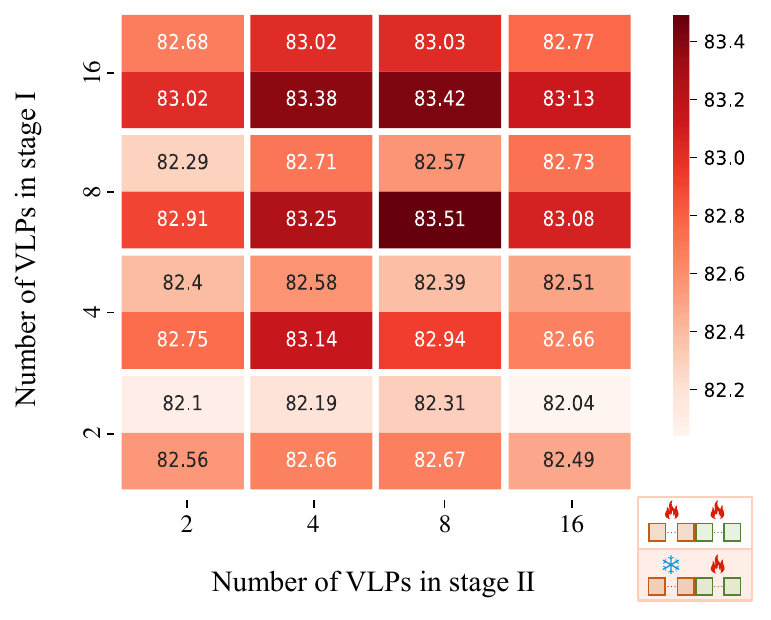}
	\caption{Ablation study on the number of Vision-Language Prompts (VLPs) in different phases and whether the boosting prompts (i.e.,\ \boostbox) are learnable in the second phase. 
 }
	\label{fig:ablation-length}
\end{figure}

\begin{table}[t]
    \centering
    \caption{Accuracy comparisons by aligning the equivalent number of VLPs. \boostbox \ denotes frozen boosting prompt and \adaptbox \ \ denotes learnable adatping prompt. \important{CasPL significantly improves the performance under the same VLP tokens in total.}}
    \setlength{\tabcolsep}{10pt}
    \resizebox{0.75\linewidth}{!}
    {
        \begin{tabular}{crccccc}
        \hline
        Method & \makecell[l]{Detail} &\makecell[c]{Number} &  Base & Novel & HM \\
        \hline
        PromptSRC & 8$\times$\adaptbox & 8 &85.15 	&76.12 	&80.38  \\
        \rowcolor{tablecolor} \textbf{+CasPL} & 4$\times$\boostbox \  + 4$\times$\adaptbox &8 &86.69 	&79.86 	&83.14   \\
        \hline
        PromptSRC & 16$\times$\adaptbox &16  &85.40 &75.85 	&80.34   \\
        \rowcolor{tablecolor} \textbf{+CasPL} & 8$\times$\boostbox \ + 8$\times$\adaptbox&16   & 86.82 	&80.44 	&83.51  \\

        \hline
        \end{tabular}
    }
    \label{table:ablation-effectiveness}
\end{table}

\noindent{\textbf{Effectiveness of boosting prompts:}}
We assess the performance of PromptSRC and CasPL by introducing an equal number of additional prompts, as outlined in Table~\ref{table:ablation-effectiveness}. In general, the approach of utilizing a cascade of boosting prompts along with adapting prompts outperforms the strategy of solely relying on the same number of adapting prompts. \rewrite{Additionally, Table~\ref{table:multi-phase-analyse} shows HM scores across 11 datasets for adding different prompts. Incorporating the boosting prompt boosts CLIP scores by 6.79\%, and adding both prompts yields a higher increase (+10.99\%). These findings suggest that the domain-general knowledge extracted by the boosting prompt assists the adapting prompt in better understanding task-specific knowledge, thereby reducing the risk of overfitting.}

\begin{table}[t]
    \centering
    \caption{\rewrite{Ablation study on decoupling domain-general and task-specific knowledge extraction. This table shows the average HM results across 11 datasets for base-to-novel generalization. ``boosting prompt'' refers to CLIP adding the boosting prompt for zero-shot inference. ``adapting prompt'' denotes PromptSRC fine-tuning. ``both'' signifies PromptSRC +CasPL fine-tuning.}}
    \resizebox{1\linewidth}{!}
    {
        \begin{tabular}{c|cccccccccccc}
        \hline
        Method      & ImageNet & Caltech101 & DTD   & EuroSAT & Cars & Flowers & Aircraft & SUN397 & Pets & UCF101 & Food & Average \\
        \hline
        CLIP        & 70.22    & 95.40      & 56.37 & 60.03   & 68.65        & 74.83      & 31.09        & 72.23  & 94.12      & 73.85  & 90.66   & 71.70   \\
        \hline
        +boosting prompt      & 73.62    & 96.20      & 67.48 & 81.37   & 76.00        & 82.59      & 38.29        & 75.74  & \textbf{96.87}      & 81.54  & \textbf{92.40}   & 78.49   \\
        +adapting prompt   & 74.01    &  96.02      & 71.75 & 82.32   & 76.58        & 85.95      & 40.15        & 80.52  &  96.30      & 82.74  & 91.10   & 79.97   \\
        both   & \textbf{74.50}    & \textbf{97.13}      & \textbf{76.44} & \textbf{90.95}   & \textbf{81.11}        & \textbf{88.46}      & \textbf{44.88}        & \textbf{81.28}  & 96.78      & \textbf{85.01}  & 91.73   & \textbf{82.69}   \\
        \hline
        \end{tabular}
    }
    \label{table:multi-phase-analyse}
\end{table}

\noindent{\textbf{Significance of multi-phase:}}
CasPL accomplishes prompt training with diverse functionalities through a multi-phase training strategy.
\rewrite{In contrast to the one-stage methods, two-stage training represents a new Prompt Learning paradigm with the following key points: (1) \textbf{Decoupling of domain-general and task-specific knowledge.} The results in Table~\ref{table:ablate-utilizing-CLIPs} demonstrate that our multi-phase design exhibits clear advantages over its single-phase counterpart (+2.72\%). This indicates that our multi-phase paradigm effectively decouples and mitigates the optimization dilemma.
(2) \textbf{Plug-and-play.} Training the boosting prompt only once allows its integration into the other methods, increasing the range from 0.12\% to 7.64\% across various tasks (Table~\ref{table:base-to-novel},~\ref{table:domain-dataset} and Fig.~\ref{fig:fewshot}). 
(3) \textbf{Small model with efficient inference.} Inserting boosting prompt enhances the performance of a smaller model (PromptSRC (ViT-B/16) +CasPL) to match that of a larger model (PromptSRC (ViT-L/14)). Thus, a two-stage paradigm approach offers advantages for deploying models in settings with limited computational resources, where only smaller models are viable.}

\begin{table}[t]
    \centering
    \caption{Ablation study on the effectiveness of the multi-phase design of CasPL on 11 datasets. ``ZS'' denotes ``Zero-Shot'' learning.} 
    \setlength{\tabcolsep}{8pt}
    \resizebox{0.75\linewidth}{!}
    {
        \begin{tabular}{l|ccc|c}
        \hline
        Method                          &KD Teacher                             & ZS & Phase  & HM    \\ 
        \hline
        CLIP (ViT-B/16)                   & \multirow{2}{*}{\makecell[c]{$\times$}} & \multirow{2}{*}{\makecell[c]{$\checkmark$}} &\multirow{2}{*}{\makecell[c]{single}}   &71.70 \\
        CLIP (ViT-L/14)                   &                                   &  &   &78.77 \\
        \hline
        PromptSRC (ViT-B/16)              & \multirow{2}{*}{\makecell[c]{$\times$}} &  \multirow{2}{*}{\makecell[c]{$\times$}} &\multirow{2}{*}{\makecell[c]{single}}   & 79.97 \\
        PromptSRC (ViT-L/14)              &                                   &  &   & 83.17 \\
        \hline
        PromptSRC (ViT-B/16) +CasPL                 &CLIP (ViT-L/14)           &\makecell[c]{$\times$}  &multiple    & 82.69 \\
        \hline
        \end{tabular}
    }
    \label{table:ablate-utilizing-CLIPs}
\end{table}

\noindent{\textbf{Visualization of different methods:}} \rewrite{Fig.~\ref{fig:ablation-tsne} shows the visualization results of different methods on base to novel generalization. CLIP with boosting prompt or adapting prompt (PromptSRC) reduces intra-class distance and increases inter-class distance. When both prompts are added simultaneously (CasPL), intra-class distance decreases further, and inter-class distance rises further. These highlight the effectiveness of multi-phase decoupling domain-general and task-specific knowledge extraction.}

\begin{figure}[t]
	\centering
	\includegraphics[width=1\linewidth]{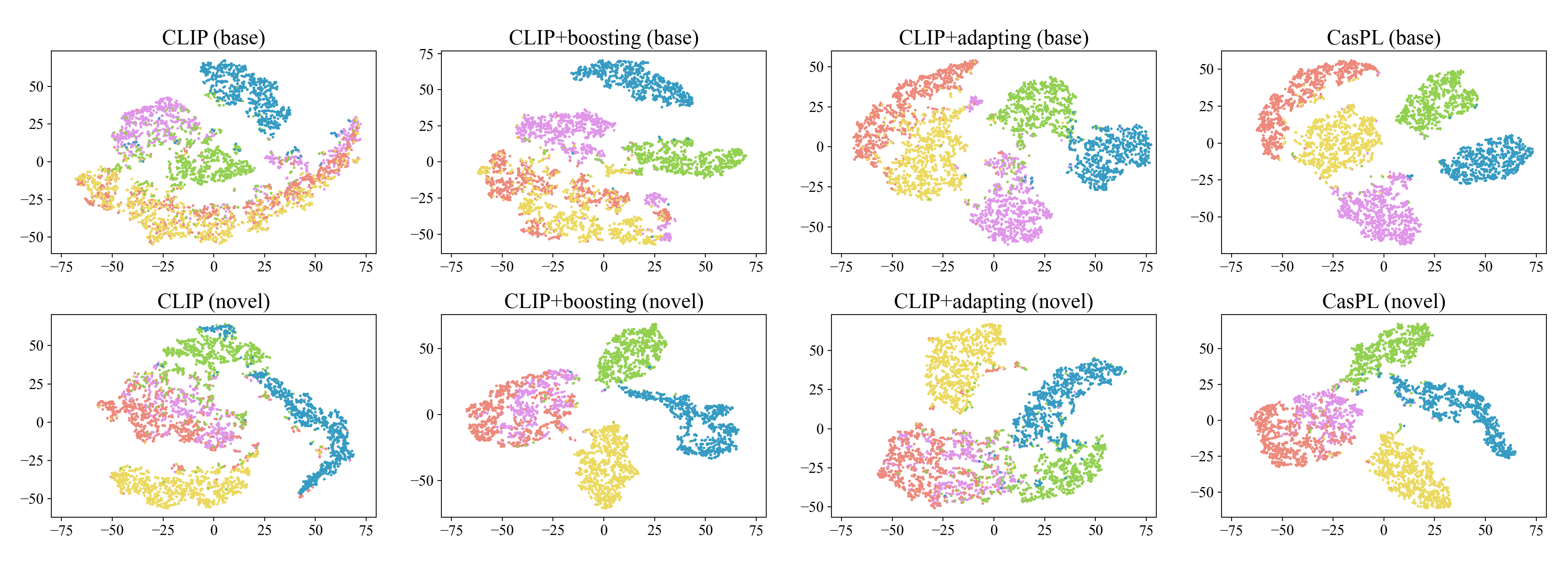}
	\caption{\rewrite{Visualization of different methods on the DTD dataset. The first row and the second row respectively depict the visualization of on base or novel categories. CasPL reduces intra-class distance and increases inter-class distance, showing its effectiveness.}}
	\label{fig:ablation-tsne}
\end{figure}

\section{Conclusion}
In this paper, we introduce the Cascade Prompt Learning (CasPL) framework, which delves into the diverse roles of prompts—specifically boosting and adapting—in vision-language models. CasPL is a new learning paradigm that introduces a two-phase training process: the first phase utilizes numerous unlabeled images to distill knowledge from the larger CLIP model, enabling boosting prompts to acquire more generalized knowledge. In the second stage, frozen boosting prompts are cascaded with newer learnable adapting prompts from existing prompt learning approaches. We comprehensively validate the effectiveness of CasPL across 11 datasets. We anticipate that our work will offer new insights into prompt learning for adapting vision-language models and facilitate the deployment of small models in resource-constrained environments.

\noindent{\textbf{Limitations and future work:}}
\implus{Our CasPL introduces the boosting prompts as plugins into existing methods with \implus{negligible inference cost and additional parameters ($< 0.1\%$).} However, it's worth noting that the boosting prompts in the first phase require training efforts in each domain, which does introduce additional computation overhead. \rewrite{In future research, our plan is to explore methodologies that leverage large-scale pre-training to enable boosting prompts to generalize better across various domain datasets and minimize additional computation time. Ideally, we aim to achieve this through a single pre-training session, eliminating the need to train individual boosting prompts for each domain.}}

\section*{Acknowledgements}
This research was supported by the Young Scientists Fund of the National Natural Science Foundation of China (Grant No.62206134), the Fundamental Research Funds for the Central Universities 070-63233084, and the Tianjin Key Laboratory of Visual Computing and Intelligent Perception (VCIP). Computation is supported by the Supercomputing Center of Nankai University (NKSC). This work was supported by the National Science Fund of China under Grant No. 62361166670.

%
%
\bibliographystyle{splncs04}
\bibliography{main}

\clearpage
\setcounter{page}{1}
\appendix
\title{Cascade Prompt Learning for Vision-Language Model Adaptation\\Supplementary Material} 

\author{Ge Wu\inst{1}$^\dag$\orcidlink{0009-0008-3011-091X},
Xin Zhang\inst{1}$^\dag$\orcidlink{0009-0000-9078-9110},
Zheng Li\inst{1}\orcidlink{0000-0003-3309-1087},
Zhaowei Chen\inst{3}\orcidlink{0000-0002-9508-6999},
\\Jiajun Liang\inst{3}\orcidlink{0000-0001-5586-340X},
Jian Yang\inst{1}\orcidlink{0000-0003-4800-832X},
Xiang Li\inst{1,2}$^*$\orcidlink{0000-0002-4996-7365}
}

\footnotetext[1]{Equal contributions. Work is done when Ge Wu is an intern at Megvii Technology.}%
\footnotetext[2]{Corresponding author.}%

\authorrunning{G. Wu et al.}

\institute{VCIP, CS, Nankai University 
\and NKIARI, Shenzhen Futian
\and Megvii Technology
\\ \email{gewu.nku@gmail.com, \{zhasion, zhengli97\}@mail.nankai.edu.cn, \\ \{csjyang, xiang.li.implus\}@nankai.edu.cn, \\ \{chenzhaowei, liangjiajun\}@megvii.com}
}

\titlerunning{Cascade Prompt Learning}

\maketitle


\section{Additional ablation studies}

\noindent{\textbf{Impact of training epoch for the first phase:}} 
Fig.~\ref{fig:abalation-epochs} (left) shows the impact of training epochs in the first stage on CasPL performance with the DTD dataset. The accuracy of the base class remains stable with increasing epochs, while the accuracy of the novel class decreases after 20 epochs.

\noindent{\textbf{Distillation temperature of learning boosting prompts:}}
The temperature hyperparameter regulates the softness of the distributions. Therefore, in Fig.~\ref{fig:abalation-epochs} right, we examine the influence of employing different temperature hyperparameters to train boosting prompts in the first stage and then fine-tuning adapter prompts in the second stage, specifically on the DTD dataset. 
According to the results, HM demonstrates the best performance when the temperature is set to 1. Hence, the temperature hyperparameter is default set to 1.

\begin{figure}[h]
	\centering
	\includegraphics[width=0.7\linewidth]{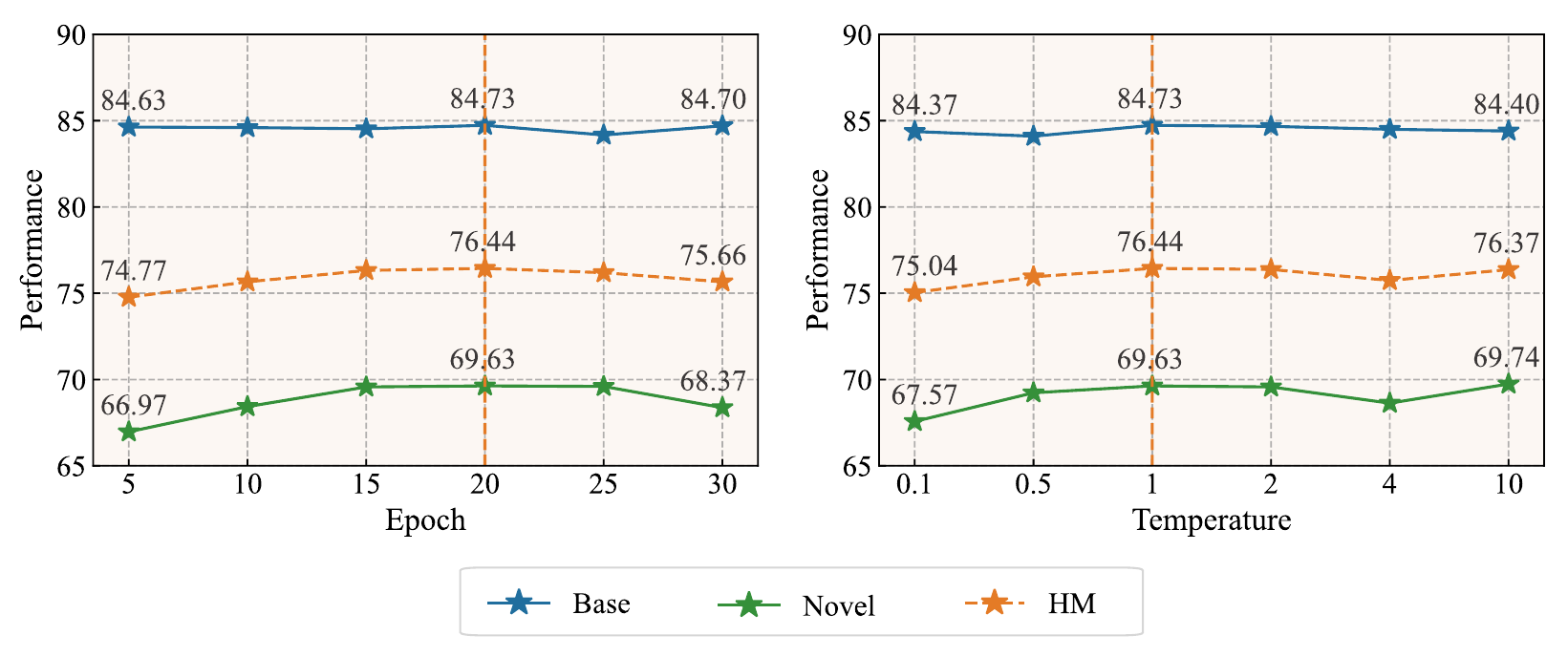}
	\caption{Ablation study on the number of training epochs for the first phase (left) and the choice of temperature hyperparameter in Eq.~\eqref{eq_kd} (right), based on the DTD dataset.}
	\label{fig:abalation-epochs}
\end{figure}

\noindent{\textbf{CLIP with boosting prompts for zero-shot inference:}}
Table~\ref{table:inference-domain} investigates the efficacy of integrating boosting prompts into CLIP for zero-shot inference. It demonstrates the accuracy improvement in domain generalization for CLIP (ViT-B/16)+ boosting prompt on both source and target datasets. However, solely using boosting prompts is less effective compared to our two-stage CasPL, as shown by the comparison with Table~\ref{table:base-to-novel}. This highlights the distinct roles played by the boosting prompts and the adapting prompts in our proposed framework.

\begin{table}[t]
    \centering
    \setlength{\tabcolsep}{10pt}
    \caption{The accuracy of zero-shot inference on domain generalization by CLIP (ViT-B/16) with adding boosting prompts. Boosting prompts can assist CLIP in enhancing domain generalization performance.}
    \resizebox{0.9\linewidth}{!}
    {
        \begin{tabular}{cllll}
        \hline
        \multirow{3}{*}{Method} & Source & \multicolumn{3}{c}{Target} \\
        \cmidrule(lr){2-2} \cmidrule(lr){3-5}
        & ImageNet & ImageNet-V2 & ImageNet-S  & ImageNet-R   \\
        \hline\noalign{\smallskip}
        CLIP & 66.73 & 60.83 & 46.15  & 73.96   \\
        \rowcolor{tablecolor}\textbf{+ boosting} & 70.40 \improve{(+3.67)}   &63.30 \improve{(+2.47)}   &47.70 \improve{(+1.55)}    &75.30 \improve{(+1.34)}     \\
        \hline
        \end{tabular}
    }
    \label{table:inference-domain}
\end{table}

\noindent{\textbf{Unsupervised training of boosting prompts using partial data:}}
This section investigates the impact of training boosting prompts with varying quantities of data on the outcomes of CasPL. Table~\ref{table:boost-few-shot} presents the DTD dataset's corresponding  HM values for different quantities. It is observed that, with an increase in the number of instances per category, the performance metric exhibits an overall upward trend, and PromptSRC +CasPL outperforms best through training on the entire dataset. Notably, when the instances per class are four or more, the HM of PromptSRC +CasPL ($\ge$72.31$\%$) exceeds that of PromptSRC HM (71.75$\%$), underscoring the effectiveness of boosting prompts.


\begin{table}[t]
    \centering
    \setlength{\tabcolsep}{10pt}
    \caption{\zhangxin{Ablation study on the HM results of boosting prompts trained with varying amounts of unlabeled images per class from the DTD dataset. (``Full'' indicates the utilization of the entire unlabeled dataset.) Utilizing more unlabeled data enables the boosting prompt to acquire more domain-general knowledge.}}
    \resizebox{0.85\linewidth}{!}
    {
        \begin{tabular}{c|lllllll}
            \hline
            Number & 1     & 2     & 4     & 8     & 16    & 32    & Full                      \\
            \hline
            HM     & 62.68 & 70.40 & 72.31 & 74.35 & 74.91 & 75.40 & \multicolumn{1}{r}{76.44} \\
            \hline
        \end{tabular}
    }
    \label{table:boost-few-shot}
\end{table}

\section{Additional implementation details}
\subsection{Boosting prompt phase}
\noindent{\textbf{General training details:}}
For the first phase of CasPL, we train the boosting prompts with a layer depth of 12, prompt length of 8, and a learning rate of 0.0025 using the SGD optimizer for 20 epochs. All learnable prompts are initialized with a normal distribution. \zhangxin{To streamline the training of the boosting prompts on ImageNet, we utilize 8 NVIDIA 3090 GPUs, while all other experiments are conducted on a single NVIDIA 3090.}

\noindent{\textbf{Text templates for senior teacher CLIP}}
Drawing from previous findings~\cite{khattak2023self}, we utilize diverse prompt templates tailored to different datasets, aiming to augment the senior CLIP's text representation ability and enhance the boosting prompts' distillation effect. Table~\ref{table:templates} presents the templates for each dataset.

\begin{table}[t]
    \centering
    \setlength{\tabcolsep}{15pt}    
    \caption{Text template utilized by senior CLIP teacher for different datasets.}
    \resizebox{0.7\linewidth}{!}
    {
        \begin{tabular}{cl}
        \hline\noalign{\smallskip}
        \makecell[c]{Dataset} & \makecell[c]{Text template}\\
        \hline
        OxfordPets    & \textsf{`` a photo of a [class], a type of pet. "}      \\
        Flowers102    & \textsf{`` a photo of a [class], a type of flower. "}      \\
        Food101       & \textsf{`` a photo of [class], a type of food. "}     \\
        FGVC Aircraft & \textsf{`` a photo of a [class], a type of aircraft. "}      \\
        DTD           & \textsf{`` [class] texture. "}      \\
        EuroSAT       & \textsf{`` a centered satellite photo of [class]. "}     \\
        UCF101        & \textsf{`` a photo of a person doing [class]. "}      \\
        other datasets      & \textsf{`` a photo of a [class]. "}  \\
        \hline
        \end{tabular}
    }
    \label{table:templates}
\end{table}

\begin{figure*}[t]
	\centering
	\includegraphics[width=1\linewidth]{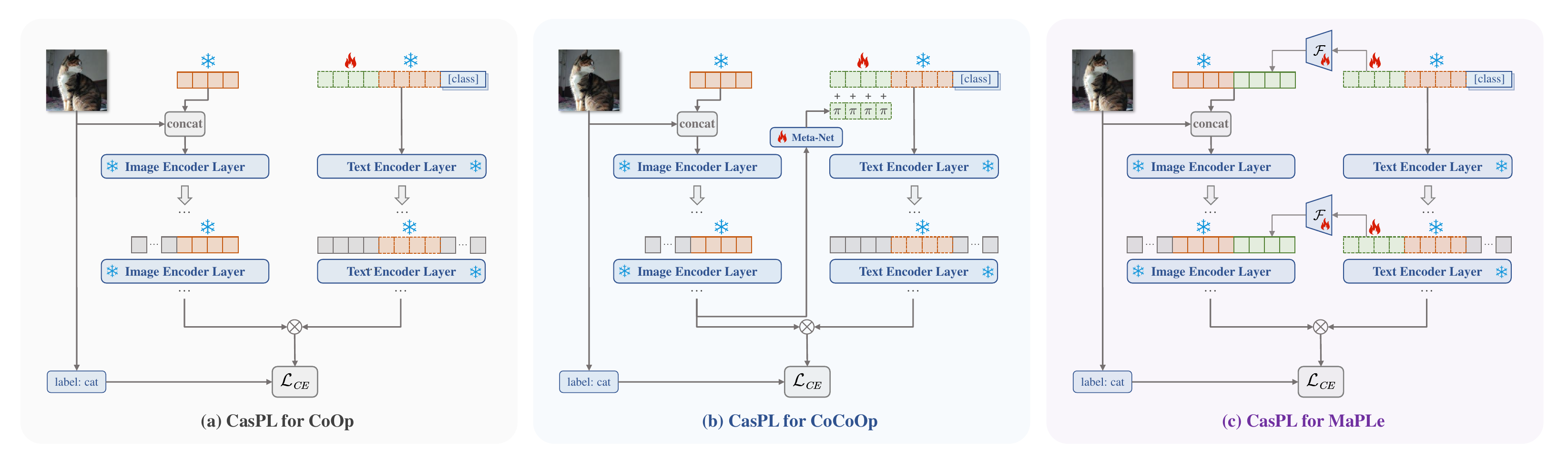}
	\caption{The detail of CasPL for previous methods. \indexbf{(a)} CoOp~\cite{zhou2022learning} employs multiple layers of text-image boosting prompts and a single layer of text adapting prompts. \indexbf{(b)} CoCoOp~\cite{zhou2022conditional} utilizes multiple layers of text-image boosting prompts and a single layer of modal blending adapting prompts. \indexbf{(c)} MaPLe~\cite{khattak2023maple} uses multiple layers of text-image boosting prompts and multiple layers of multi-modal adapting prompts.}
	\label{fig:methods-detail}
\end{figure*}

\subsection{Adapting prompt phase}
\noindent{\textbf{Base-to-Novel generalization:}}
The training details of each of the previous methods on this task are shown in Table~\ref{table:training-setting}.
Various prior approaches based on prompt learning exhibit differences in the specifics of their implementation.
CoOp~\cite{zhou2022learning} employs single learnable text prompts (see Fig.~\ref{fig:methods-detail} (a)). CoCoOp~\cite{zhou2022conditional} combines image features with learnable text prompts to obtain multi-modal information (see Fig.~\ref{fig:methods-detail} (b)).  MaPLe~\cite{khattak2023maple} utilizes multi-layer learnable text prompts and image prompts generated from the text prompts (see Fig.~\ref{fig:methods-detail} (c)).  Specific details of PromptSRC are elaborated in Fig.~\ref{fig:framework}.

\noindent{\textbf{Domain generalization:}}
Following the previous method in this task~\cite{khattak2023maple}, we adjust the parameters for MaPLe, specifically setting its optimizer's learning rate to 0.0026 and establishing the training epoch at 2. 

\noindent{\textbf{Few-shot experiments:}}
Following the methodology from previous work~\cite{khattak2023self}, we set the training epoch for PromptSRC at 50, keeping the other configurations consistent with those outlined in Table~\ref{table:training-setting}. The main text features comparison curves, while additional numerical results are available in Table~\ref{tab_appendix:few_shot_experiments}.

\noindent{\textbf{Compare with un-/weakly-supervised methods:}}
In this experiment, the CLIP zero-shot method utilizes simple templates as the text input, and the numerical results are derived from the official UPL~\cite{huang2022unsupervised} code.
To ensure a fair comparison, the three training strategies in ENCLIP~\cite{menghini2023enhancing} are implemented based on the PromptSRC pipeline~\cite{khattak2023self}. Few-pseudo labels (FPL) utilizes 16 pseudo labels per novel class and 16 labeled data per base class. Iterative Refinement of FPL (IFPL) utilizes the same training data as FPL but involves multiple iterations. The labels are recalculated in each iteration, and the prompt is reinitialized. Grow and Refine Iteratively Pseudolabels (GRIP) gradually increases the number of unlabeled datasets compared to IFPL (with a maximum limit of 16 per class in our implementation).

\begin{table}[t]
    \centering
    \setlength{\tabcolsep}{15pt}
    \caption{Training settings for base-to-novel generalization task.}
    \resizebox{0.9\linewidth}{!}
    {
        \begin{tabular}{ccccc}
        \hline\noalign{\smallskip}
        ~& CoOp & CoCoOp & MaPLe & PromptSRC \\
        \hline
        Vision Prompt Length & - & - & 8 & 8 \\ 
        Text Prompt Length   & 8 & 8 & 8 & 8 \\ 
        Prompt Layer         & 1 & 1 & 12 & 12 \\ 
        Optimizer            & SGD & SGD & SGD & SGD \\ 
        Learning Rate        & 0.002 & 0.002 & 0.0035 & 0.0025 \\ 
        Epoch                & 50 & 10 & 5 & 20 \\
        \hline
        \end{tabular}
    }
    \label{table:training-setting}
\end{table}

\begin{table*}[t!]
    \small \centering
    \setlength{\tabcolsep}{15pt}
    \caption{The performance of CasPL (built on PromptSRC) compared to other methods in the few-shot setting. Results across various few-shot setups demonstrate CasPL's ability to enhance model performance.}
    \scalebox{0.63}[0.63]{
        \begin{tabular}{ll|ccccc}
        \hline
        Dataset & 
        Method & 
        1 shot &
        2 shots & 
        4 shots & 
        8 shots &
        16 shots \\  
        \hline
        \multirow{5}{*}{ImageNet}      & Linear probe CLIP      &32.13	&44.88&	54.85	&62.23	&67.31\\ 
                                       & CoOp                   &66.33	&67.07	&68.73	&70.63&	71.87\\ 
                                       & CoCoOp                  & 69.43	&69.78	&70.39&	70.63&	70.83\\
                                       & MaPLe                  & 62.67	&65.10	&67.70&	70.30&	72.33\\
                                       & PromptSRC             & 68.13	&69.77	&71.07	&72.33	&73.17\\
                                       \rowcolor{tablecolor}
                                       & CasPL \textbf{(Ours) }     &68.73   &70.07   &71.43    &72.87   & 74.20    \\
        \hline
        \multirow{5}{*}{Caltech101}    & Linear probe CLIP      & 79.88	&89.01	&92.05	&93.41	&95.43\\
                                       & CoOp                    & 92.60	&93.07	&94.40	&94.37	&95.57\\
                                       & CoCoOp                  & 93.83	&94.82	&94.98	&95.04	&95.16\\
                                       & MaPLe                  & 92.57	&93.97	&94.43&	95.20&	96.00\\
                                       &  PromptSRC              &93.67	&94.53	&95.27	&95.67	&96.07\\
                                       \rowcolor{tablecolor}
                                       & CasPL \textbf{(Ours) }    &93.97 	&95.20 	&96.10 	&96.23 	&96.80   \\
                                       \hline
        \multirow{5}{*}{DTD}           & Linear probe CLIP       & 34.59	&40.76	&55.71	&63.46	&69.96\\
                                       & CoOp                   &50.23&	53.60	&58.70	&64.77	&69.87\\
                                       & CoCoOp                      & 48.54	&52.17	&55.04	&58.89	&63.04\\
                                       & MaPLe                  & 52.13	&55.50	&61.00&	66.50&	71.33\\
                                       &  PromptSRC                  & 56.23&59.97	&65.53	&69.87	&72.73\\
                                       \rowcolor{tablecolor}
                                        & CasPL \textbf{(Ours) }    &62.63  &63.67 &69.07 &71.00 	&75.13      \\
                                       \hline
        \multirow{5}{*}{EuroSAT}       & Linear probe CLIP       &49.23	&61.98	&77.09&	84.43	&87.21\\
                                       & CoOp                   &54.93	&65.17	&70.80	&78.07	&84.93\\
                                       & CoCoOp                  & 55.33	&46.74	&65.56&	68.21	&73.32\\
                                        & MaPLe                  & 71.80	&78.30	&84.50&	87.73&	92.33\\
                                       & PromptSRC              &73.13 &79.37	&86.30	&88.80	&92.43\\
                                       \rowcolor{tablecolor}
                                        & CasPL \textbf{(Ours) }   &83.40 &86.53 	&91.07 	&91.07 	&94.17     \\
        \hline
        \multirow{5}{*}{StanfordCars}  & Linear probe CLIP     & 35.66	&50.28	&63.38	&73.67	&80.44\\
                                   & CoOp                        &67.43	&70.50	&74.47	&79.30	&83.07\\
                                           & CoCoOp              & 67.22	&68.37	&69.39	&70.44&	71.57\\
                                         & MaPLe                  & 66.60	&71.60	&75.30&	79.47&	83.57\\
                                      &  PromptSRC               &69.40	&73.40	&77.13	&80.97	&83.83\\
                                      \rowcolor{tablecolor}
                                       & CasPL \textbf{(Ours) }    &72.80 	&77.23 	&80.03 	&83.30 	&86.70     \\
        \hline
        \multirow{5}{*}{Flowers102}    & Linear probe CLIP      & 69.74	&85.07	&92.02	&96.10	&97.37\\
                                       & CoOp                   & 77.53	&87.33	&92.17	&94.97	&97.07\\
                                       & CoCoOp                    &72.08	&75.79	&78.40	&84.30	&87.84\\
                                       & MaPLe                  & 83.30	&88.93	&92.67&	95.80&	97.00\\
                                       & PromptSRC            & 85.93	&91.17	&93.87	&96.27	&97.60\\
                                       \rowcolor{tablecolor}
                                        & CasPL \textbf{(Ours) }  &90.33 &94.17  &95.53  &97.20  &98.30     \\
                                       \hline
        \multirow{5}{*}{FGVCAircraft}  & Linear probe CLIP       & 19.61	&26.41	&32.33	&39.35&	45.36\\
                                       & CoOp                     & 21.37&	26.20	&30.83	&39.00	&43.40\\
                                       & CoCoOp                   & 12.68	&15.06	&24.79	&26.61	&31.21\\
                                       & MaPLe                  & 26.73	&30.90	& 34.87&	42.00&	48.40\\
                                       & PromptSRC           & 27.67	&31.70	&37.47&	43.27	&50.83\\
                                       \rowcolor{tablecolor}
                                         & CasPL \textbf{(Ours) }   &32.80 &35.20 	&41.03 	&48.03 	&55.37  \\
                                       \hline
        \multirow{5}{*}{SUN397}        & Linear probe CLIP          & 41.58	&53.70	&63.00	&69.08	&73.28\\
                                       & CoOp                     & 66.77	&66.53	&69.97	&71.53	&74.67\\
                                       & CoCoOp                   & 68.33	&69.03&	70.21	&70.84	&72.15\\
                                       & MaPLe                  & 64.77	&67.10	&70.67&	73.23&	75.53\\
                                      & PromptSRC                &69.67	&71.60	&74.00	&75.73	&77.23\\
                                      \rowcolor{tablecolor}
                                       & CasPL \textbf{(Ours) }   & 71.03 	&72.70 	&74.53 	&76.33 &77.70      \\
                                       \hline
        \multirow{5}{*}{OxfordPets}    & Linear probe CLIP        & 44.06	&58.37	&71.17	&78.36	&85.34\\
                                       & CoOp                      & 90.37	&89.80	&92.57	&91.27	&91.87\\
                                       & CoCoOp                       & 91.27	&92.64	&92.81	&93.45	&93.34\\
                                       & MaPLe                  & 89.10	&90.87	&91.90&	92.57& 92.83\\
                                       & PromptSRC                   & 92.00	&92.50	&93.43	&93.50	&93.67\\
                                       \rowcolor{tablecolor}
                                        & CasPL \textbf{(Ours) }    & 92.97 	&93.37 	&93.97 &93.93 &94.13     \\
                                       \hline
        \multirow{5}{*}{UCF101}        & Linear probe CLIP        & 53.66	&65.78	&73.28	&79.34&	82.11\\
                                       & CoOp                     & 71.23	&73.43	&77.10	&80.20	&82.23\\
                                       & CoCoOp                   & 70.30	&73.51	&74.82	&77.14&	78.14\\
                                       & MaPLe                  & 71.83	&74.60	& 78.47& 81.37&	85.03\\
                                       & PromptSRC                   & 74.80	&78.50	&81.57	&84.30	&86.47\\
                                       \rowcolor{tablecolor}
                                        & CasPL \textbf{(Ours) }     &79.53 	&82.03 	&84.77 	&86.70 	&88.47      \\
                                       \hline
        \multirow{5}{*}{Food101}       & Linear probe CLIP       & 43.96	&61.51	&73.19	&79.79	&82.90\\
                                       & CoOp                    & 84.33	&84.40	&84.47	&82.67	&84.20\\
                                       & CoCoOp                     & 85.65	&86.22	&86.88	&86.97	&87.25\\
                                       & MaPLe                  & 80.50 &81.47	&81.77&	83.60&	85.33\\
                                       & PromptSRC                & 84.87&	85.70	&86.17	&86.90	&87.5\\
                                       \rowcolor{tablecolor}
                                        & CasPL \textbf{(Ours) }   &86.80 &87.20 	&87.40 	&87.80 	&88.40      \\
                                        \hline
        \multirow{5}{*}{Average}       & Linear probe CLIP       & 45.83	&57.98	&68.01	&74.47&	78.79\\
                                       & CoOp                    & 67.56	&70.65	&74.02	&76.98	&79.89\\
                                       & CoCoOp                & 66.79&	67.65	&71.21&	72.96	&74.90\\
                                       & MaPLe                  & 69.27	&72.58	&75.37&	78.89&	81.79\\
                                       & PromptSRC               & 72.32	&75.29	&78.35	&80.69	&82.87\\
                                       \rowcolor{tablecolor} 
                                       & CasPL \textbf{(Ours) } &  75.91 & 77.94  & 80.45& 82.22 & 84.49   \\
        
        \hline
        \end{tabular}
        }
    \label{tab_appendix:few_shot_experiments}
\end{table*}



\end{document}